\documentclass[runningheads]{llncs}

\usepackage[ruled, linesnumbered, noline]{algorithm2e}
\usepackage[dvipsnames]{xcolor}
\usepackage{soul}

\usepackage{graphicx}
\begin{document}
\title{Image Augmentation for Multitask Few-Shot Learning: \\
Agricultural Domain Use-Case}
%
%


\author{Sergey Nesteruk \and
Dmitrii Shadrin \and
Mariia Pukalchik
}
%
%
\institute{
Skolkovo Institute of Science and Technology, Digital Agriculture Lab
\email{sergei.nesteruk@skoltech.ru}}

\maketitle 
\begin{abstract}
Large datasets' availability is catalyzing a rapid expansion of deep learning in general and computer vision in particular. At the same time, in many domains, a sufficient amount of training data is lacking, which may become an obstacle to the practical application of computer vision techniques. This paper challenges small and imbalanced datasets based on the example of a plant phenomics domain. We introduce an image augmentation framework, which enables us to extremely enlarge the number of training samples while providing the data for such tasks as object detection, semantic segmentation, instance segmentation, object counting, image denoising, and classification. We prove that our augmentation method increases model performance when only a few training samples are available. In our experiment, we use the DeepLabV3 model on semantic segmentation task with \textit{Arabidopsis} and \textit{Nicotiana tabacum} image dataset. The obtained result shows a 9\% relative increase in model performance compared to the basic image augmentation techniques.

\end{abstract}

\section{Introduction}

Machine learning and computer vision algorithms have recently shown the capabilities to address various challenging industrial and scientific problems. Successful application of machine learning and computer vision algorithms for solving complex tasks is impossible without relying on comprehensive and high-quality training and testing data~\cite{kwon2013effects},~\cite{sbai2020impact},~\cite{zendel2017good}. Deep learning algorithms, in particular, computer vision algorithms for solving classification, object detection, semantic and instance segmentation problems require a huge variety of input data (images) for ensuring future robust work of the trained models~\cite{barbedo2018impact},~\cite{zheng2016improving},~\cite{hendrycks2020many}. There are two ways for enhancing the characteristics of the used for training purposes datasets. The first one is obvious and implies the physical collection of the dataset samples in various conditions to ensure the high diversity of the training data. There is set of huge datasets that were collected at the beginning of the rapid development of the deep learning approaches for solving computer vision problems. These datasets are commonly used as the benchmark~\cite{deng2009imagenet},~\cite{lin2014microsoft},~\cite{xia2018dota},~\cite{caba2015activitynet}. One of the specifics of these datasets is that they are general and, first of all, suitable for comparison of the power of the developed algorithms. However, presented labeled data can be almost useless for solving specific industrial problems. One of the possible applications of such well-known datasets is that they can serve as a good basis for pre-training of neural networks (transfer learning)~\cite{tan2018survey}. Using these pre-trained neural networks, it is possible to fine-tune them and adapt to address specific problems. However, in some cases, even for the fine-tuning, a comprehensive dataset is in high demand. Some events are rare, and it is possible to collect only a few data samples~\cite{vannucci2016classification},~\cite{wang2020generalizing}. Thus, the second approach for enhancing the characteristics of the dataset can help. This approach is based on the artificial manipulations with the initial dataset. One of the well-developed techniques is data augmentation, where initial images transformed according to special rules~\cite{shorten2019survey}.

The agricultural domain is one of the industrial and research areas for which the development of artificial methods for improvement of the training datasets is vital~\cite{Kuznichov2019AugmentationCounting},~\cite{fawakherji2020data},~\cite{wu2020dcgan}. This demand appears due to the high complexity and variability of the investigated system (plant) that has to be characterized by computer vision algorithms. There is a huge amount of different plant species, and plants grow slowly. Thus, collecting and labeling huge datasets for each specific plant growing in each specific stage is quite a complex task. Overall, it can be outlined that it is difficult to collect datasets, especially of plants, and expensive to annotate them~\cite{Ching2018Opportunities_biology}. Therefore we propose a method to multiply the number of training samples. It does not require many computational resources and can be performed on the fly. 

This paper is organized as follows: in the rest of this section, we overview image datasets and image augmentation techniques in the agricultural domain; in section~\ref{sec:approach}, we describe our approach to image augmentation; in section~\ref{sec:experiments}, we show the advantage of the proposed approach for few-short learning on the example of semantic segmentation problem. 


The notation that we use in this paper is listed in Table~\ref{tab:notation}.

\begin{table}
\begin{center}
\caption{Notations.}
\label{tab:notation}
\begin{tabular}{|l|l|}
\hline
{\bfseries Notation} &  {\bfseries Description} \\
\hline
n & The number of objects per scene \\
\hline

m & The number of output masks \\
\hline

p & Average packaging overhead per input object \\
\hline

o & Average overhead for auxiliary data storage per object \\
\hline

\`{o} & Constant system overhead \\
\hline

s & Objects' shrinkage ratio \\
\hline

$\theta$ & Orientation coefficient \\
\hline

H & The set of objects heights \\
\hline

$\widetilde{H}$ & The set of shrinked objects heights \\ 
\hline

W & The set of objects heights \\
\hline

$\widetilde{W}$ & The set of shrinked objects widths \\
\hline

$\overline{h}$ & Average over all input object heights \\ 
\hline

$\overline{w}$ & Average over all input object widths \\
\hline

$\hat{h}$ & Hard height restriction \\
\hline

M & Average RAM~(Random Access Memory) usage \\
\hline
\end{tabular}
\end{center}
\end{table}

\subsubsection{Plant Image Datasets.}

Publicly available plant image datasets are crucial in precision agriculture as they reduce the time and effort spent on data collection and preparation. Also, more data enable more efficient algorithms to be developed and evaluated to address various computer vision problems.

According to research~\cite{LU2020105760} currently, there are 34 popular public agricultural image datasets, among which 15 datasets are intended for weed control, 10 datasets for fruit detection, and the remaining 9 datasets for other applications. The existing datasets are not universal enough since the specific problem can rather strongly determine them, have a rather small scale (less than 1000 images per class), or not quite diverse (include only some growth stages, lighting and weather conditions). In addition to the small number of samples and lack of universality, many have a rather mediocre annotation quality. 

For example, the IPPN dataset~\cite{MINERVINI2016IPPNdataset} contains mistakes in segmentation annotation -- some of the leaves coming from the picture corners are not annotated, which leads to misleading accuracy scores. Besides, this dataset's backgrounds are not consistent enough, leading to incorrect functioning of algorithms trained on this data set in field conditions.

\subsubsection{Plant Image Synthesis and Augmentation.}

Computer vision models require much training data. Therefore it becomes challenging to obtain a good model with limited datasets. Namely, a small-capacity model might not capture complex patterns, while a big capacity model tends to overfit if small datasets are used~\cite{Feng_2019_CVPR_Workshops}.

To overcome this issue, we use various image augmentation techniques. Data augmentation aims to add diversity to the train set and to complicate the task for a model~\cite{Zeiler2014Visualizing}. Among plant image augmentation approaches we can distinguish: basic computer vision augmentations, learned augmentation, graphical modeling, augmentation policy learning, collaging, and compositions of the ones above.

\textbf{Basic computer vision augmentations} are the default methods preventing overfitting in most computer vision tasks. They include image cropping, scaling, flipping, rotating, and adding noise~\cite{Krizhevsky2017ImageNet}. There are also advanced augmentation methods, connected with distortion techniques and coordinate system changes~\cite{buslaev2018albumentations}. Since these operations are quite generic, most popular ML frameworks support them. However, though being helpful, these methods demonstrate a limited use  as they bring  insufficient diversity to the training data.

\textbf{Learned augmentation} stands for generating training samples with an ML model. For this purpose, conditional Generative Adversarial Networks~(cGANs) and Variational Autoencoders~(VAEs) are frequently used. In the agricultural domain, there are examples of applying GANs to \textit{Arabidopsis} plants images for the leaf counting task~\cite{zhu2018data},~\cite{Giuffrida_2017_ICCV}. The main drawback of this approach is that generating an image with a neural network is quite resource-intensive. Another disadvantage is the overall pipeline complexity: the errors of a model that generates training samples are cumulated with the errors of a model that solves the target task.

\textbf{Graphical modeling} is another popular method in plant phenomics. It involves creating a 3D model of a plant and rendering it. The  advantage of this  process  is  that it permits to generate large  datasets  with  precise  annotations,  as the labels of each pixel are known.However, this technique is highly resource-intensive; moreover, the results obtained using the existing solutions~\cite{BARTH2018Data_synthesis},~\cite{Daniel2018Segmentation_Synthetic} seem to look quite artificial.

\textbf{Learned augmentation policy} is a series of techniques used to find a combinations of basic augmentations that maximize model generalization. This implies hard binding of the learned policy to the ML model, the dataset, and the task. Although it is shown to provide systematic generalization improvement on object detection~\cite{Barret2019Augmentation} and classification~\cite{Lemley2017Smart_Augmentation}, its universal character as well as the ability to be performed along with multi-task learning are not supported with solid evidence.

\textbf{Collaging} presupposes cropping an object from an input image with the help of a manually annotated mask, and pasting it to a new background with basic augmentations of each object~\cite{Kuznichov2019AugmentationCounting}.

All the above-listed methods can be characterised by the same disadvantage. In existing implementations, they are not suitable for multiple task learning. For many computer vision applications, we have to train models to find multiple types of masks, perform object counting and object detection. In particular plant phenomics case, it is required to solve semantic segmentation, plant detection, leaf counting, and other tasks. Even if it is not needed to solve these tasks directly for our goal, it is usually beneficial to pre-train a model on auxiliary tasks to reach a higher score. It is vital to have a tool that generates various types of realistic training samples using a limited amount of input data. Therefore in the next sections, we introduce and validate a lightweight, comprehensive model-agnostic method of image augmentation designed for solving various computer vision tasks.

\section{Our Approach}
\label{sec:approach}

In this paper, we introduce a method of image augmentation for segmentation tasks. Our method takes image-mask pairs and transforms them to obtain various scenes. Having a set of image-mask pairs, we can place multiple of them on a random-chosen background. Transformation of input data and background, accompanied by adding noise gives possibility for us to synthesize an infinite number of compound scenes.

We distinguish between several types of image masks:

\begin{itemize}
  \item \textbf{Single (\textit{S})} -- single-channel mask that shows the object presence.
  \item \textbf{Multi-object (\textit{MO})} -- multi-channel mask with a special color for each object (for each plant).
  \item \textbf{Multi-part (\textit{MP})} -- multi-channel mask with a special color for each object part (for each leaf).
  \item \textbf{Semantic (\textit{Sema})} -- multi-channel mask with a special color for each type of object (leaf, root, flower).
  \item \textbf{Class (\textit{C})} -- multi-channel mask with a special color for each class (plant variety).
\end{itemize}

\begin{table}[!h]
\begin{center}
\caption{Possible mask transitions.}
\label{tab:mask_type}
\begin{tabular}{|c|p{0.1\textwidth}|p{0.1\textwidth}|p{0.1\textwidth}|p{0.1\textwidth}|p{0.1\textwidth}|}
\hline
{\bfseries Input mask type} &  S & MO & MP & Sema & C \\
\hline

S & + & + & - & - & + \\
\hline

MP & + & + & + & - & + \\
\hline

Sema & + & + & - & + & + \\
\hline

\end{tabular}
\end{center}
\end{table}

Noteworthy, a single input mask type allows us to produce more than one output mask type. Hence, multiple tasks can be solved using any  dataset, even the one that was not originally designed for these tasks (see Table~\ref{tab:mask_type} for the possible mask transitions).

For example, an image with a multipart mask as input enables us to produce: the \textit{S} mask, which is a boolean representation of any other mask, the \textit{MO} mask with unique colors for every object, the \textit{MP} mask with a unique color for each part across all the present objects, and the \textit{C} mask that distinguishes the classes. Additionally, for every generated sample, we provide bounding boxes for all objects and the number of objects of each class. 

Note that we assume that each input image-mask pair includes a single object. Therefore we can produce the \textit{MO} mask based on any other mask. To create the \textit{C} mask, the information about input objects must be provided.

\subsection{System Architecture}

\begin{figure}

\includegraphics[width=\textwidth]{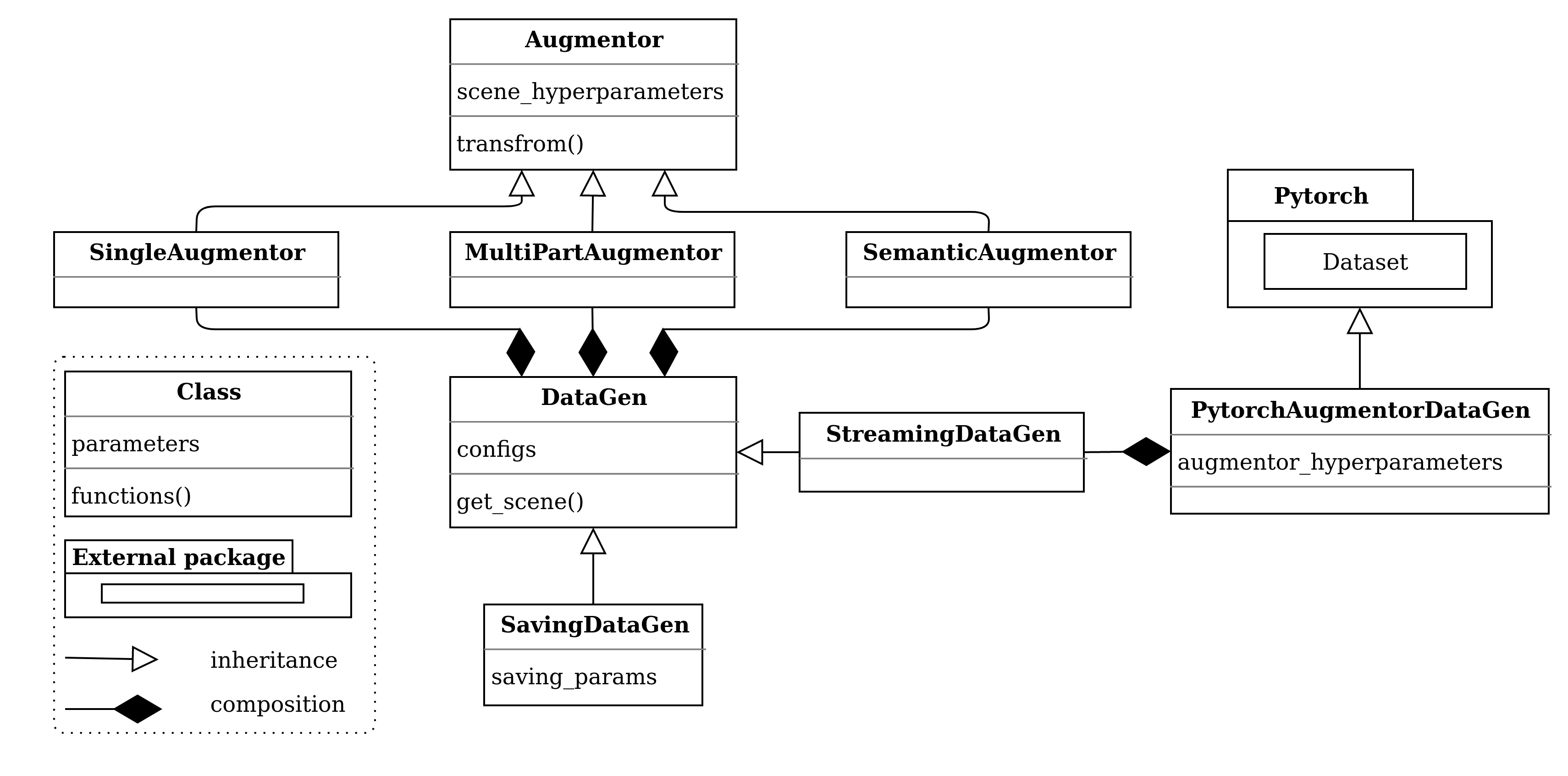}
\caption{Augmentation System Architecture.} \label{fig:architecture}
\end{figure}

The overall code structure is presented in Figure~\ref{fig:architecture}. \textit{The library with the code will be shared as an open source code with the community.} The core of the presented system is the \textit{Augmentor}. This class implements all the image and mask transformations. Such transformations as flipping or rotating are mutual for both image and mask. We add noise for images only. 

From the main \textit{Augmentor} class we inherit \textit{SingleAugmentor}, \textit{MultiPartAugmentor} and \textit{SemanticAugmentor} classes allowing to apply different input mask types and to treat them separately. To be more precise, \textit{SingleAugmentor} is exploited for \textit{S} input mask type, \textit{MultiPartAugmentor} is for \textit{MP} mask type and \textit{SemanticAugmentor} is for \textit{Sema} mask type.

The described above classes are used in the \textit{DataGen} class which chooses images for each scene and balances classes if needed. Two principal ways of new scenes generation are offline and online. We implement them in \textit{SavingDataGen} and \textit{StreamingDataGen} accordingly. Both of the classes take the path to images with corresponding masks as input. The offline data generator produces a new folder with created scenes while the online generator can be used to load data directly to a neural network. 

The offline generation is more time-consuming  because of additional disk access operations; at the same time, it is performed in advance and thus does not affect model training time. It also makes it easier to manually look through the obtained samples to tune transformation parameters.

Meanwhile, the online data generator streams its results immediately to the model without saving images on the disk. What is more, this type of generator allows us to change parameters on the fly: for instance,  training the model gets started on easy samples, and then  the complexity may be manipulated based on the loss function.

\subsection{Implementation Details}

\begin{figure}[!h]
\includegraphics[width=\textwidth]{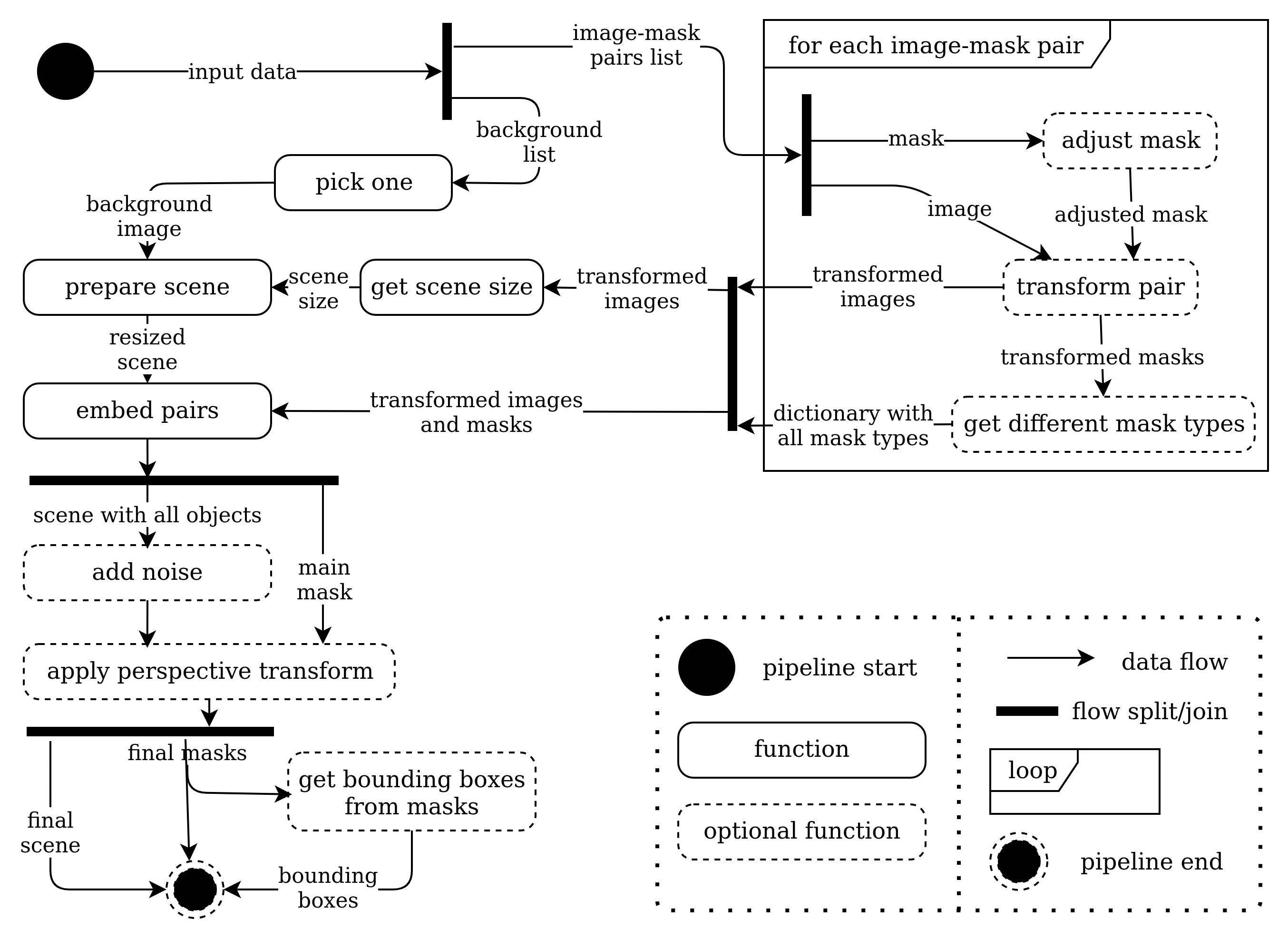}
\caption{Transformation Pipeline Activity Diagram.} \label{fig:activity}
\end{figure}

The present section discusses the main transformation pipeline (Figure~\ref{fig:activity}).

The first step is to select the required number of image-mask pairs from a dataset. By default, we pick objects with repetitions that enable us to create scenes with a larger number of objects than present in input data. 

After that, we prepare images and masks before combining them into a single scene. The procedure is following:

\begin{itemize}
  \item adjusting the masks to exclude large margins;
  \item performing the same random transformations to both image and mask;
  \item obtaining all required masks types and auxiliary data.
\end{itemize}

Once all the transformations are performed and we know the sizes of all objects, the size of the output scene is calculated. Note that input objects can have different sizes and orientations; therefore, we cannot simply place objects by grid because it will lead to inefficient space usage. It is also not a good idea to place objects randomly in most cases because it will lead to uncontrollable overlapping of objects.

Within the framework of our approach the objects are packed using the Maximal Rectangles Best Long Side Fit~(MAXRECTS-BLSF) algorithm. It is a greedy algorithm that is aimed to pack rectangles of different sizes into a bin using the smallest possible area. The maximum theoretical packaging space overhead of the MAXRECTS-BLSF algorithm is $0.087$.
The BLSF modification of the algorithm tries to avoid a significant difference between sides lengths. However, just like other rectangular packing algorithms, this one also tends to abuse the height dimension of the output scene yielding  a column-oriented result. 

In order to control both overlapping of the objects and the orientation of output scenes, we introduce two modifications to the MAXRECTS-BLSF algorithm.

\textbf{Control of the overlapping} is achieved via the substitution of objects' real sizes with the shrinked ones when passing them to the packing algorithm. The height and width are modified according to Eq.~(\ref{eq:shift_h}):

\begin{equation}
\widetilde{H} = (1-s)H; \\
\widetilde{W} = (1-s)W,
\label{eq:shift_h}
\end{equation}
where \textit{s} ranges from 0 to 1 inclusively. 



\begin{figure}
\centering

\begin{tabular}{cc}
\includegraphics[width=0.61\textwidth]{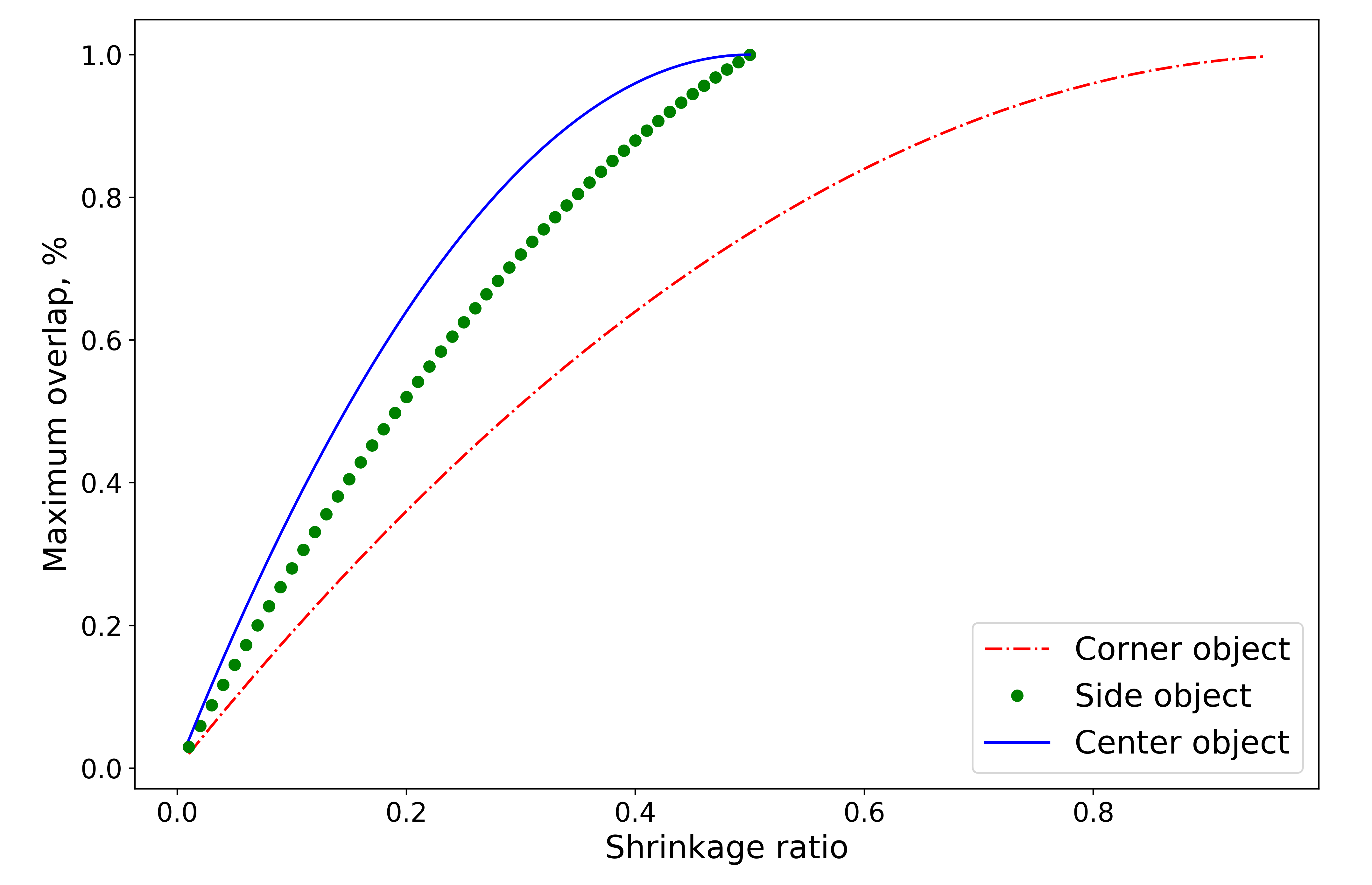} & \includegraphics[width=0.39\textwidth]{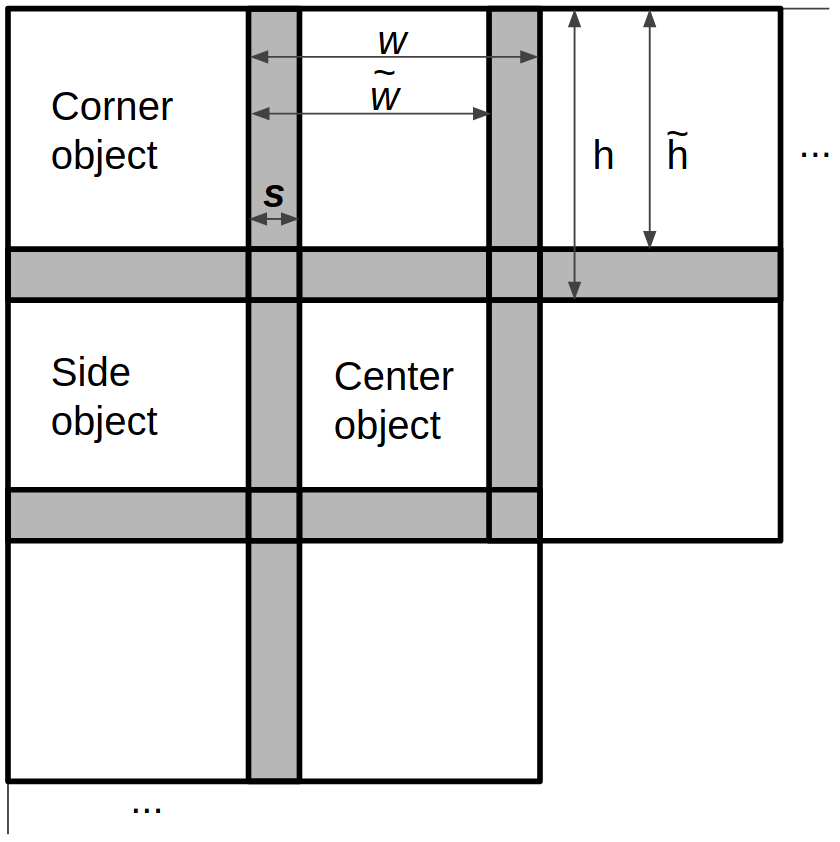}\\
(a) & (b)\\
\end{tabular}
\caption{Shrinkage ratio effect illustration (a) The dependency of maximum objects' overlap on the shrinkage ratio (b) Simplified scene generation example} 
\label{fig:overlap}
\end{figure}

The bigger the shrinkage ratio, the smaller are the substituted images. It is to be highlighted that it is applied to both height and width and to all of the input objects. The real overlapping area in practice will vary depending on each objects' shape and position. To perceive the overlap percentage see Fig.~\ref{fig:overlap}. Here we consider the case where all input objects are squares without any holes. In other words, it is the maximum possible overlap percentage for the defined shrinkage ratio. We show this value for an object in the corner of a scene, an object in the side, and an object in the middle separately.


We recommend choosing \textit{s} between $0$ and $0.3$; however, taking into consideration sparse input masks, it can be slightly higher.

\textbf{To control the orientation} of the output scene we set the hard limit of the scene height for the packing algorithm. Assuming that input objects will have different sizes in practice, we cannot get an optimal packing with the fixed output image size or width to height ratio. To calculate the hard height limit we use Eq.~(\ref{eq:height}).

\begin{equation}
 \hat{h} = max \Bigg( max H, \theta \frac{ \sum_{i=1}^{n} \widetilde{H}_{i} } { \big \lceil \sqrt{n} \big \rceil } \Bigg)
 \label{eq:height}
\end{equation}

The fraction in Eq.~(\ref{eq:height}) estimates the required value of height to make a square scene. We choose maximum between it and the biggest objects' height to ensure that it is enough space for any input object. 
The orientation coefficient $\theta$ can be treated as the target width to height ratio. It will not produce the scenes with the fixed ratio, but with many samples, the average value will approach the target one. $\theta = 1$ will try to get square scenes. $\theta > 1$ will generate landscape scenes. In our experiments we set $\theta$ to 1.2 to obtain close to square images with landscape preference. The average resulted width to height ratio over ten thousand samples was 1.1955.

To adjust the background image size to the obtained scene size, we resize the background if it is smaller than the scene or randomly crop it if it is bigger.

We generate the required number of colors, excluding black and white, and find their Cartesian product according to Algorithm~\ref{alg:colors} for coloring \textit{MO} and \textit{MP} masks. 

\begin{algorithm}[H]
\SetAlgoNoLine
\label{alg:colors}

\SetKwInOut{Input}{Input}
\SetKwInOut{Output}{Output}
    
\Input{Number of objects $n$;}
\Output{The set of colors $C$;}

$L = \lceil \sqrt[3]{n+2} \rceil$ \\
$s = \frac{1}{L}$

\For{$l=0,...,L-1$} {
    $T \gets 1-(s*l)$  
}

\Return $C = \{ (c_1, c_2, c_3) | c_1, c_2, c_3 \in T \}$

\caption{Colors generation.}
\end{algorithm}

To preserve the correspondence between the input objects and their representation on the final scene, we color the objects in order of their occurrence.

\subsection{Time Performance}

\begin{figure}[!h]
\centering
\begin{tabular}{cc}
\includegraphics[width=0.5\textwidth]{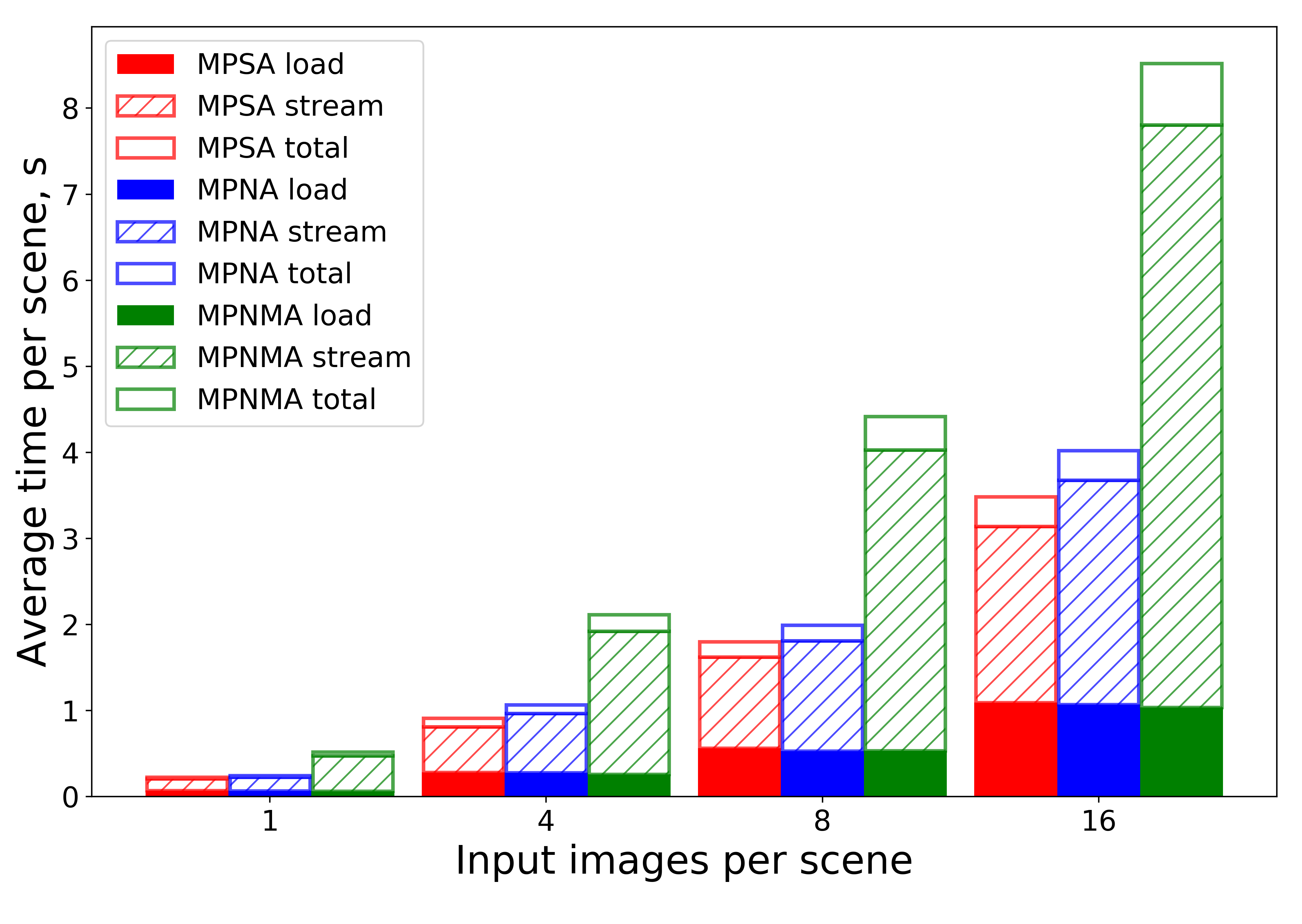} & \includegraphics[width=0.5\textwidth]{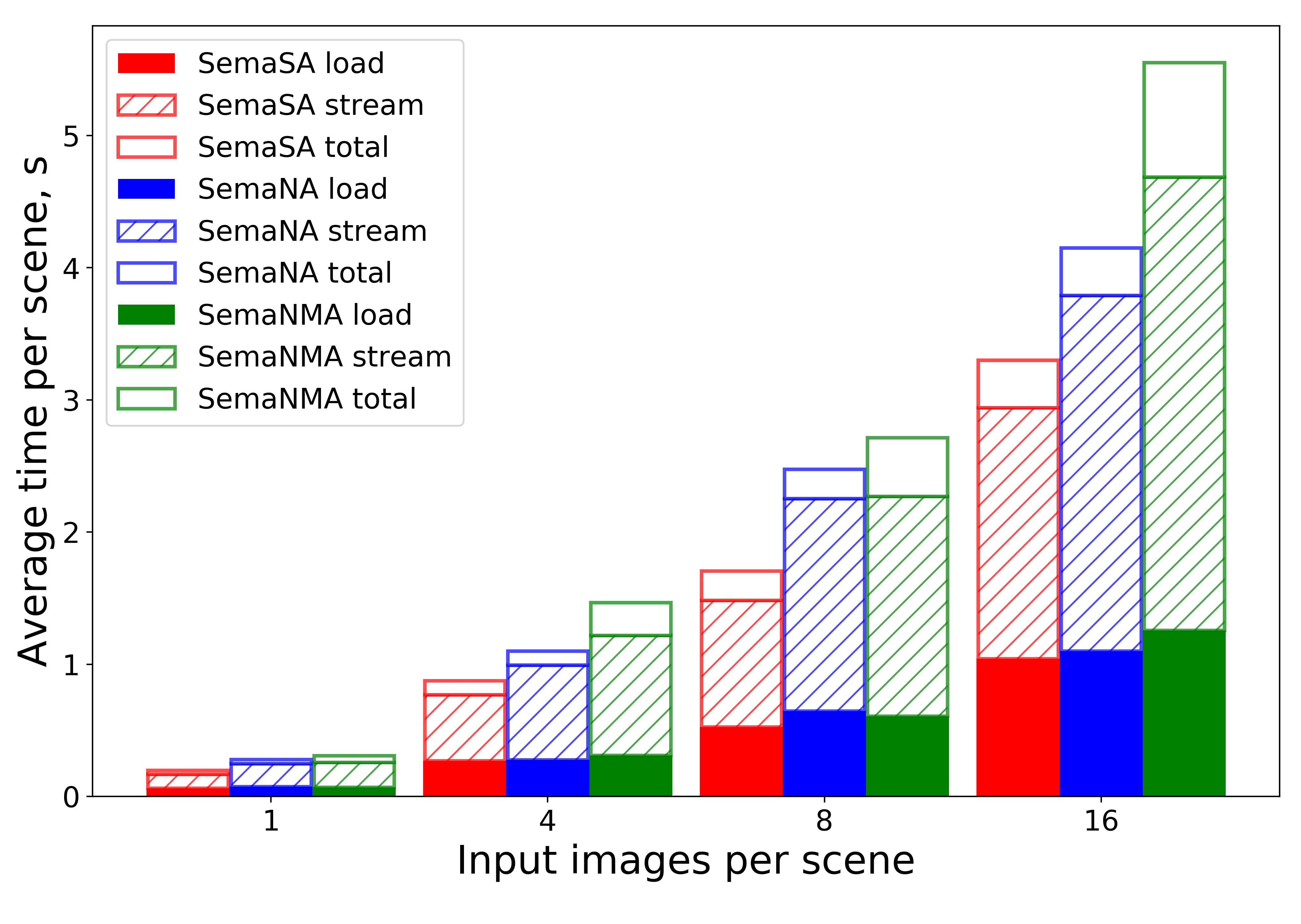}\\
(a) & (b)\\
\end{tabular}
\caption{Average scene generating time with (a) \textit{MultiPartAugmentor} and (b) \textit{SemanticAugmentor}} 
\label{fig:time}
\end{figure}

In this section, we measure the average time that is required to generate scenes of various complexity. For this experiment we use Intel(R) Core(TM) i7-7700HQ CPU 2.80~GHz without multiprocessing. The average height of objects in the dataset is 385 pixels; the average width is 390 pixels. The results are averaged on a thousand scenes for each parameter combination and are reflected in Fig.~\ref{fig:time}a for \textit{MultiPartAugmentor} and Fig.~\ref{fig:time}b for \textit{SemanticAugmentor}.

\textit{SA}~(the red bar on the left) stands for \textit{Simple Augmentor} with one type of output mask; \textit{NA}~(the blue bar in center) means adding noise and smoothing to scenes; \textit{NMA}~(the green bar on the right) means adding noise, smoothing, calculating bounding boxes, and generating all possible types of output masks. To recall possible mask types for each augmentor refer to Table~\ref{tab:mask_type}. The filled area in the bottom shows the time for loading input images and masks from disks. The shaded area in the middle shows the time for actual transformation. The empty are in the top shows the time for  saving all the results to disk. If you accumulate every bar with all the bars below it, the top of the shaded bar will show the time for \textit{StreamingDataGen}, and the top of the empty area will show the time for \textit{SavingDataGen}.


From the bar plots, you can see linear dependence between the number of input objects and the time for generating a scene.

\subsection{System Parameters}

Two main classes of the system where we can choose parameters are \textit{Augmentor} and \textit{DataGen}, or classes inherited.

The \textit{Augmentor} parameters that define the transformations are shown in Table~\ref{tab:transforms}.

\begin{table}
\centering
\caption{Augmentor transformation parameters.}
\label{tab:transforms}
\begin{tabular}{|l|l|c|c|}
\hline
\textbf{Operation} & \textbf{Description} & \textbf{Range} & \textbf{Default value}  \\
\hline

Shrinkage ratio & See Fig.~\ref{fig:overlap} for details & $[0...1)$ & 0 \\
\hline

Rotation & The maximum angle of image and mask rotation & $[0...180]$ & 180 \\
\hline

Flip probability & The probability to flip image and mask horizontally & $[0...1]$ & 0.5 \\
\hline

Salt & The probability for each pixel to be colored white & $[0...1]$ & 0 \\
\hline

Pepper & The probability for each pixel to be colored black & $[0...1]$ & 0 \\
\hline

Smooth & The size of Gauss kernel applied for image smoothing & $1, 3, ...$ & 1 \\
\hline

Perspective transform & The share of added width before perspective transform & $[0...3]$ & 0 \\
\hline

Noise & The variance of Gaussian noise & $0...$ & 0 \\
\hline

\end{tabular}
\end{table}

The rest of the \textit{Augmentor} parameters define output mask types, bounding box presence, and mask preprocessing steps.

The data generator parameters define the rules to pick samples for scenes: the number of samples per scene, picking samples for a single scene from the same class or randomly, class balancing rule, the input file structure, the output file structure.

\subsection{Limitations}

The proposed image augmentation scheme can be used when we have masks for input images. The system can work with instance segmentation masks and semantic segmentation masks. 

The system's primary usage involves generated complex scenes from simple input data; however, the scene can include a single object if needed. The key feature of the system is its ability to generate a huge amount of training samples even for the task for which the original dataset was not designed. For instance, having only an image and a multi-part mask as an input, we can produce samples for instance segmentation, instance parts segmentation, object detection, object counting, denoising, and classification. The described system can also be beneficial for few-shot learning when the original dataset is minimal.

To apply the proposed augmentation scheme successfully, the dataset should not be exceedingly sensitive to scene geometry, since such a behavior can be undesirable in some cases. For example, if you use the dataset of people or cars, the described approach by default can place one object on top of the other. Nevertheless, we can add some extra height limitations or use perspective transformation in these cases.

Another point is that we should find appropriate background images that would fit some particular case or, on the contrary, refuse using any background.

It also should be noted that the time for scene generation is close to linear when we have enough memory to store all objects and overhead for a scene. To estimate the average required RAM per scene, we use Eq.~(\ref{eq:memory})

\begin{equation}
M = 3n \overline{h} \overline{w} [ (1 + m) p + o + 2 ] + `{o} 
\label{eq:memory}
\end{equation}

In this equation, we can neglect the overhead $`{o} < o << M $, because it is considerably smaller than the data itself. 


\section{Experiments and Results}
\label{sec:experiments}


\begin{figure}[!h]
\includegraphics[width=\textwidth]{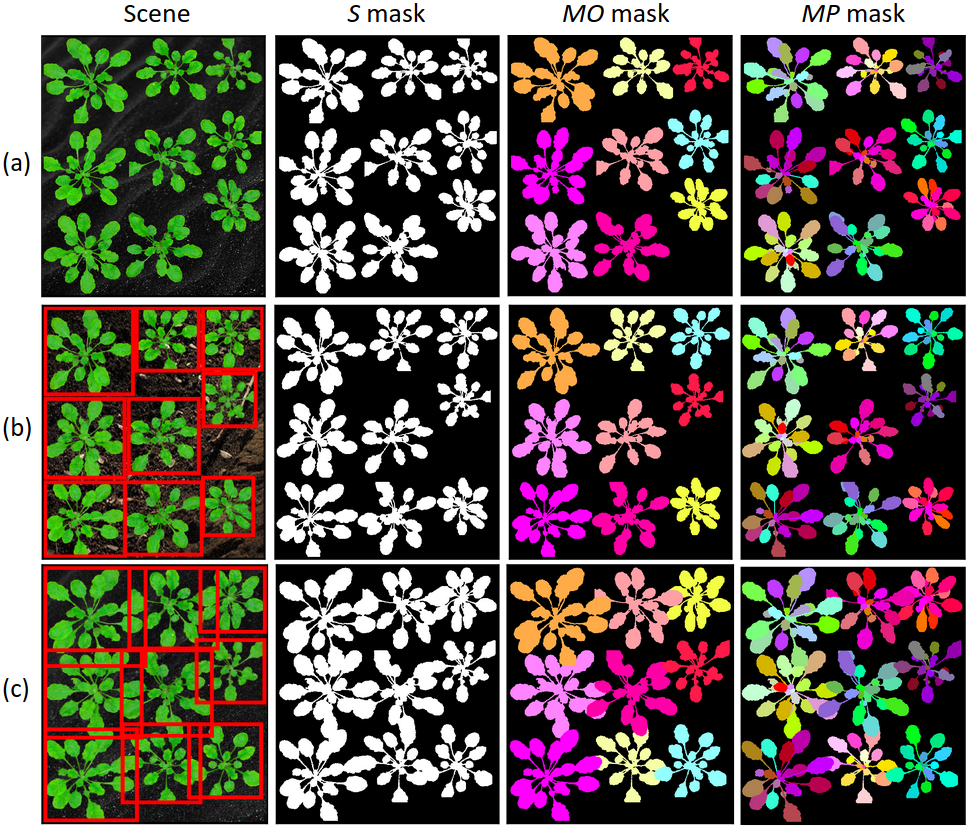}
\caption{A MultiPartAugmentor generated scene. 
(a) Without noise. 
(b) With added noise, blurring, and bounding boxes. 
(c) With added noise, blurring, bounding boxes, and $s=0.1$.
} 
\label{fig:generated_samples}
\end{figure}

This section provides some basic examples of scenes that can be produced using the proposed image augmentation system.

In the first series of examples we choose nine random samples from the dataset with possible repetitions and apply different transformation parameters to them. The background images were randomly picked from three options. Fig.~\ref{fig:generated_samples} is the result of using \textit{MultiPartAugmentor}: (a) without adding neither noise nor overlapping; (b) with noise and smoothing, and automatically generated bounding boxes for every object; (c) with objects overlapping with $s=0.1$. In \textit{S} mask type, white color means any object's presence, while the black color is for the background. In \textit{Sema} masks, the colors remain as in input masks. 








To test the performance of our framework, we use the IPPN dataset~\cite{MINERVINI2016IPPNdataset}. This dataset contains top-down view of \textit{Arabidopsis} and \textit{Nicoticana tobacum}. The original dataset contains images and segmentation masks manually labeled by experts, such that each leaf has its own label. There are 120 labeled images of \textit{Arabidopsis} and 62 samples of \textit{Nicoticana tobacum}.

\begin{figure}[!h]
\includegraphics[width=\textwidth]{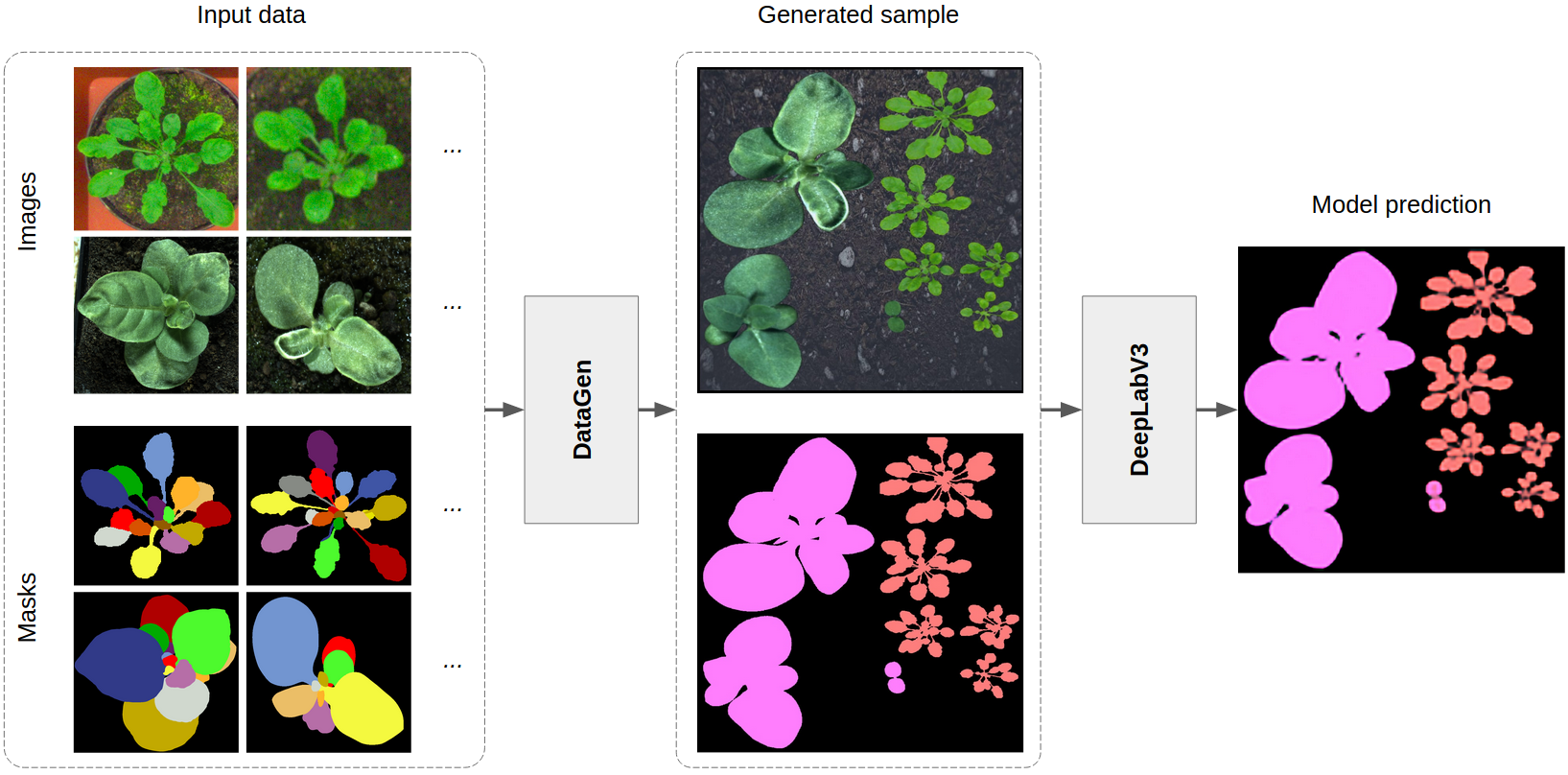}
\caption{Images data flow scheme.} \label{fig:dataflow}
\end{figure}

In our experiment, we examine the possibility of using our augmentation framework for few-shot learning. For this purpose, we use the cross-validation technique. More precisely, we select eight random image-mask pairs from both \textit{Arabidopsis} and \textit{Nicoticana tobacum} classes and use them to generate training scenes with the corresponding masks~(Fig.~\ref{fig:dataflow}). The rest of the input data is used to validate the results of the trained model. The whole pipeline is repeated several times, and the results are averaged. For every subset we generate 250 scenes with default parameters that can be seen in Tab.~\ref{tab:transforms}.

As a segmentation model we use the Pytorch implementation of DeepLabV3 model with the ResNet-101 backbone~\cite{chen2017rethinking}. We use F1 score (aka Dice score) as our key metric for results evaluation~(Eq.~\ref{eq:f1formula}). Before measuring the F1 we also apply sigmoid function to the net outputs.
\begin{equation}
\ F\textit{1} = \frac{2*TP}{2*TP+FP+FN} 
\label{eq:f1formula}
\end{equation}

To train our models on the original data we use $batch\_size = 2$, \textit{Adam} optimizer with starting learning rate of $0.0005$ for basic data augmentation and $0.0001$ for data with no augmentation, because this values showed the best results in experiments and train for 150 epochs. We also use \textit{ReduceLROnPlateau} scheduler, which reduce learning rate when a metric (loss in our case) has stopped improving. For the models trained on generated scenes we use the same parameters, but change the learning rate to $0.001$ and train for only 20 epochs, which is enough to reach best results. Loss function is defined as BCE loss with logits. This is the modified binary cross-entropy loss, which is combined with a sigmoid layer~\cite{jadon2020survey}.

For \textit{standard} augmentation we use the following methods from \textit{Albumentations} package~\cite{buslaev2018albumentations}: \textit{HorizontalFlip}, \textit{VerticalFlip}, \textit{ShiftScaleRotate}, \textit{GridDistortion}.



Experiments show, that our advanced augmentation algorithm is indeed better than standard augmentation techniques or no augmentation at all (Table~\ref{table:results}).

\setlength{\tabcolsep}{3pt}
\begin{table}
\caption{\label{table:results} Table of balanced F1 score for different augmentation techniques}
\begin{center}
\begin{tabular}{|c|c|}
\hline
\textbf{Augmentation technique}
& \textbf{Balanced F1} \\
\hline
No augmentation
& $66.18 \pm 1.57$ \% \\
Standard augmentation
& $68.21 \pm 1.52$ \% \\
Ours
& \textbf{$\textbf{74.35} \pm 2.32$} \% \\
\hline
\end{tabular}
\end{center}
\end{table}


\section{Conclusions}

In this paper, we introduce a novel image augmentation technique for few-shot learning. The presented framework allows for generating large training datasets using only a few input samples. It also provides training data for such tasks as instance segmentation, semantic segmentation, classification, object detection, and object counting, even if the original dataset was not designed for solving these problems. To show our method's advantage, we compare model performance without any augmentation, with basic augmentations, and with our augmentation on semantic segmentation task. We report a 9\% relative increase in F1-score using the DeepLabV3 model on the IPPN dataset compared to basic augmentations and a 12\% relative increase in F1-score in comparison with no augmentation.

\bibliographystyle{splncs04}
\bibliography{ref.bib}

%





\end{document}